\documentclass[journal]{IEEEtran}
\usepackage[cmex10]{amsmath}
\usepackage{amsfonts}
\usepackage{algorithmic}
\usepackage{algorithm}
\usepackage{graphicx}
\usepackage{array}
\usepackage{multirow}
\usepackage{cite}
\usepackage{url}
\usepackage[switch]{lineno}

\usepackage{bm}
\usepackage{color}
\begin{document}
\title{RGB-T Image Saliency Detection via\\ Collaborative Graph Learning}

%\author{Michael~Shell,~\IEEEmembership{Member,~IEEE,}
%        John~Doe,~\IEEEmembership{Fellow,~OSA,}
%        and~Jane~Doe,~\IEEEmembership{Life~Fellow,~IEEE}% <-this % stops a space
%\thanks{M. Shell was with the Department
%of Electrical and Computer Engineering, Georgia Institute of Technology, Atlanta,

\author{Zhengzheng~Tu, Tian~Xia, Chenglong~Li, Xiaoxiao~Wang, Yan~Ma and Jin Tang
 \thanks{T. Tu, T. Xia, C. Li, X. Wang, Y. Ma and J. Tang are with Key Lab of Intelligent Computing and Signal Processing of Ministry of Education, School of Computer Science and Technology, Anhui University, Hefei, China, Email: zhengzhengahu@163.com, tianxia.ahu@foxmail.com, lcl1314@foxmail.com, xiaoxiao9212@foxmail.com, m17856174397@163.com, tangjin@ahu.edu.cn. C. Li is also with Institute of Physical Science and Information Technology, Anhui University, Hefei, China. (\emph{Corresponding author is Chenglong Li})}
\thanks{This research is jointly supported by the National Natural Science Foundation of China (No. 61602006, 61702002, 61872005), Natural Science Foundation of Anhui Province (1808085QF187), Natural Science Foundation of Anhui Higher Education Institution of China (KJ2017A017), Open fund for Discipline Construction, Institute of Physical Science and Information Technology, Anhui University.
}}

\markboth{IEEE Transactions on Multimedia}%
{Shell \MakeLowercase{\textit{et al.}}: Bare Demo of IEEEtran.cls for IEEE Journals}
\maketitle

\begin{abstract}
Image saliency detection is an active research topic in the community of computer vision and multimedia. Fusing complementary RGB and thermal infrared data has been proven to be effective for image saliency detection. In this paper, we propose an effective approach for RGB-T image saliency detection. Our approach relies on a novel collaborative graph learning algorithm. In particular, we take superpixels as graph nodes, and collaboratively use hierarchical deep features to jointly learn graph affinity and node saliency in a unified optimization framework. Moreover, we contribute a more challenging dataset for the purpose of RGB-T image saliency detection, which contains 1000 spatially aligned RGB-T image pairs and their ground truth annotations. Extensive experiments on the public dataset and the newly created dataset suggest that the proposed approach performs favorably against the state-of-the-art RGB-T saliency detection methods.
\end{abstract}


\section{Introduction}
\label{sec::introduction}
\begin{figure}[tb]
  \centering
  \includegraphics[width=0.35\textwidth]{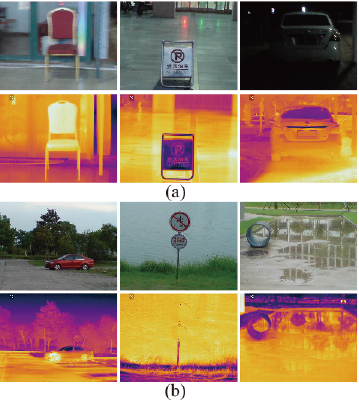}
\caption{Typically complementary advantages of RGB and thermal data.
(a) Advantages of thermal data over RGB data, where visible spectrum is influenced by blur, reflective light and low illumination. (b) Advantages of RGB data over thermal data, where thermal spectrum is influenced by thermal reflection and thermal crossover.}
\label{fig::RGBTcompare}
\end{figure}

\IEEEPARstart{T}{he} goal of image saliency detection is to estimate visually most salient and important objects in a scene, and has wide applications in the community of computer vision and multimedia. In the past decade, image saliency detection has been extensively studied, but still faces many challenges in adverse environments. Integrating visible and thermal infrared (RGB-T) data has proven to be effective for several computer vision tasks~\cite{li2016learning,li18arxiv,li2017weighted,Li18eccv,Li2017A}. Thermal infrared cameras can capture infrared radiation emitted by the object whose temperature is above absolute zero, and thus are insensitive to illumination variation and have a strong ability to penetrate haze and smog, as shown in Fig.~\ref{fig::RGBTcompare}.

RGB-T image saliency detection is relatively new in the computer vision community, and there are few methods to work on it. As the initial advance, Li et al.~\cite{Li2017A} propose a multi-task manifold ranking algorithm for RGB-T image saliency detection, and built up a unified RGB-T image benchmark dataset. Although this work achieves a significant step in the aspect of RGB-T saliency detection, the performance might be limited by the following issues: i) The handcraft features are only adopted to compute saliency values. ii) The graph structure is fixed which only considers the local neighbors, and not able to capture more intrinsic relationships among graph nodes. iii) The graph construction and the saliency computation are independent phases.

To handle these problems, we propose a novel approach for RGB-T image saliency detection, and formulate RGB-T image saliency detection as a graph learning problem. First, we segment input RGB and thermal images jointly into a set of superpixels. Since the deeper layers contain richer semantic information to localize salient regions while the shallower layers have much finer structures to retain clearly object boundaries~\cite{Liu2016DHSNet,HouCvpr2017Dss,ZhangAJLL18,Xiao2018Deep}, we extract multi-level deep features~\cite{Long2015Fully} from these superpixels for each modality. For each modality and layer, we construct a graph with superpixels as nodes, and each node is connected to its neighbors with an edge, whose weight indicates the appearance compatibility of two neighboring nodes. These graphs only take local relationships into account and their structures are fixed so that the intrinsic relationships among superpixels are not utilized~\cite{Li17aaai,Li18pami}. To better depict the intrinsic relationships among superpixels and capture saliency cues, we adaptively utilize the affinity matrices calculated in multiple kinds of feature spaces to learn the collaborative graph. In particular, we jointly learn the collaborative graph including graph structure, edge weights indicating appearance compatibility of two neighboring nodes and node weights representing saliency values in a single unified optimization framework.

In addition, existing RGB-T image benchmark dataset for saliency detection~\cite{Li2017A} has several limitations: i) The alignment errors might be large. The used RGB and thermal cameras have totally different imaging parameters and are mounted on tripods, and they use a homography matrix to approximate the transformations of two images. ii) The aligned method introduces blank boundaries in some modality, which might destroy the boundary prior to some extent. iii) Most of scenes are very simple, which makes the dataset less challenge and diverse. In this paper, we contribute a larger dataset for the purpose of RGB-T image saliency detection. The imaging hardware includes highly aligned RGB and thermal cameras, and the transformation between two modal images are thus only translation and scale. This setup makes the images of different modalities highly aligned, and have no blank boundaries. Furthermore, we take more challenges and diversities into account when building up the dataset and collect 1000 RGB-T image pairs.

The major contributions of this work are summarized as follow:\\

\begin{itemize}

\item We propose a novel graph learning approach for RGB-T image saliency detection. Extensive experiments on both public and newly created dataset against the state-of-the-art methods demonstrate the effectiveness of the proposed approach.

\item We present a new optimization algorithm to jointly learn graph structure, edge weights indicating appearance compatibility of two neighboring nodes and node weights representing saliency values in a single unified framework.

\item We create a new benchmark dataset containing 1000 aligned RGB and thermal image pairs and their ground truth annotations for performance evaluation of different RGB-T saliency detection methods. This dataset has been released to public~\footnote{RGB-T Image Saliency Detection Dataset:\\http://chenglongli.cn/people/lcl/dataset-code.html}.

\end{itemize}

\section{Related Work}

\subsection{RGB Image Saliency Detection}
Numerous image saliency models have proposed based on various cues or principles, and can be divided into the following two types: bottom-up data-driven models and top-down task-driven models.

\begin{figure}[tb]
  \centering
  \includegraphics[width=0.35\textwidth]{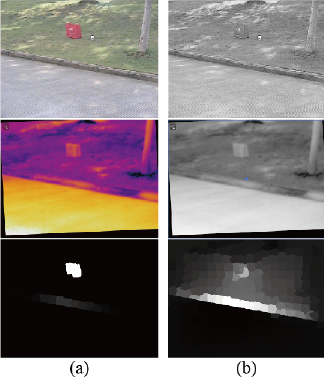}
\caption{Illustration of advantages of RGB data over grayscale ones.
(a) RGB image, thermal image and their saliency map estimated by our algorithm. (b) Grayscale image, thermal image and their saliency map estimated by our algorithm.}
\label{fig::RGB-grayscale}
\end{figure}

Bottom-up data-driven models take the underlying image features and some priors~\cite{Harel2006Graph,Gopalakrishnan10tip,Yang2013Saliency,Li2015Robust,Wang2016GraB,Dou2017Object,jiang2018saliency,Chen2018Bi,Quan2018Unsupervised} into consideration, such as color, orientation, texture, boundary, and contrast.  Then Gopalakrishnan \emph{et al.}~\cite{Gopalakrishnan10tip} performed Markov random walks on a complete graph and a k-regular graph to detect the salient object. In~\cite{Yang2013Saliency}, they employed a manifold ranking technique to detect salient objects, which performed a two-stage ranking with the background and foreground queries to generate the saliency maps. Li \emph{et al.}~\cite{Li2015Robust} formulated pixel-wised saliency maps via regularized random walks ranking, from the superpixel-based background and foreground saliency estimations. Wang \emph{et al}.~\cite{Wang2016GraB} proposed a new graph model, in which they not only adopted local and global contrast, but also enforced the background connectivity constraint, and optimized the boundary prior. And the model in~\cite{ZhangAJLL18} extracted multi-layer deep features, then constructed the sparsely connected graph to obtain the local context information of each node.

Top-down models always learn salient object detectors, recently most of which are based on deep learning networks~\cite{li2015visual,Wang2016Kernelized,Liu2016DHSNet,HouCvpr2017Dss}. Liu \emph{et al.}~\cite{Liu2016DHSNet} proposed a deep hierarchical saliency network (DHSNet) based on convolutional networks for saliency detection, then introduced hierarchical recurrent convolutional neural network (HRCNN) for refining the details of saliency maps by combining local context information. In ~\cite{HouCvpr2017Dss}, Hou \emph{et al.} introduced short connections to the skip-layer structure by transforming high-level features to shallower side-output layers and thus obtain ideal results.~\cite{li2015visual} used multi-scale features extracted from convolutional neural networks and proposed a saliency framework which integrates CNN-based model to obtain saliency map. Wang \emph{et al.}~\cite{Wang2016Kernelized} made use of object-level proposals and region-based convolutional neural network (R-CNN) features for saliency detection. In general, their performances are better than bottom-up models. However, top-down methods always need time-consuming training processes.

\begin{figure*}[t]
\centering
\includegraphics[width =0.9\textwidth]{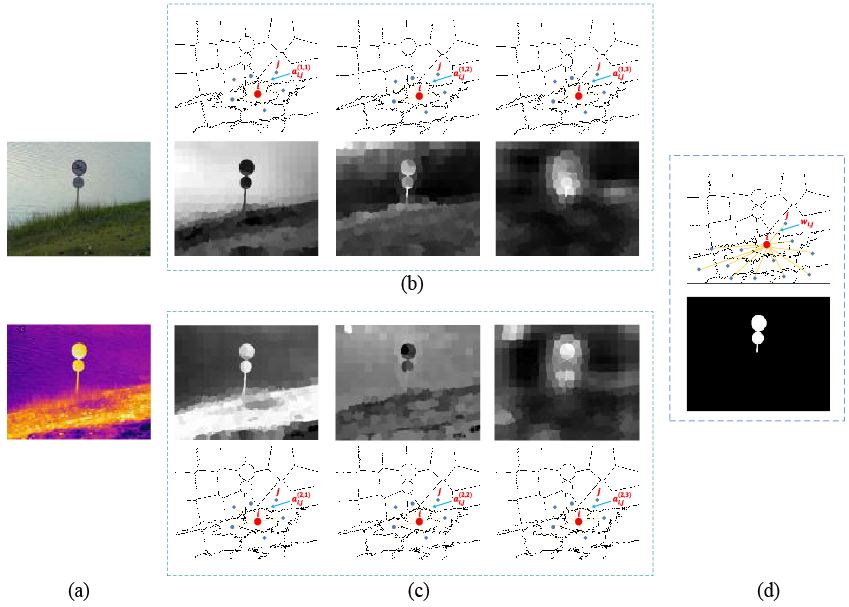}
\caption{Illustration of collaborative graph learning. (a) Input RGB and thermal images. (b) Three-layer deep features extracted from the RGB image and the corresponding graphs, where the feature map of each layer is obtained by averaging all channels for clarity. Here, only the $i$-th superpixel is shown for clarity. The local neighbors are connected with the $i$-th superpixel, and each edge is assigned with a weight, $a_{ij}^{(m,k)}$, please see our model for details. (c) Three-layer deep features extracted from the thermal image and the corresponding graphs. (d) Learnt graph, where the graph structure (i.e., node connections), edge weights (i.e., $w_{ij}$) and node weights (i.e., saliency values of superpixels) are jointly optimized using multi-layer deep features of RGB and thermal images. }\label{fig::motivation}
\end{figure*}

\subsection{RGB-T Vision Methods}

 Integrating RGB and thermal infrared data has drawn more attentions in the computer vision community~\cite{Yang2017Fast,Sun2012Fusion,Li2017Grayscale,li2017weighted,Li18eccv,Li2017A} with the popularity of thermal sensors. There are several typical problems that use these two modalities. For the problem of grayscale-thermal foreground detection, Yang \emph{et al.}~\cite{Yang2017Fast} proposed a collaborative low-rank decomposition approach to achieve cross-modality consistency, and also incorporated the modality weights to achieve adaptive fusion of multiple source data. Herein, grayscale-thermal is the special case of RGB-T, where grayscale denotes one-channel gray image. For clarity, we present an example to justify the advantages of RGB images over grayscale ones, as shown in Fig.~\ref{fig::RGB-grayscale}. In fact, RGB data provide more color information than grayscale, and thus achieve more robust results in some tasks, such as saliency detection. For example, in Fig.~\ref{fig::RGB-grayscale}, the red bag is discriminative against the green grass in RGB image, but their intensities (grayscale image) are very similar. We run our algorithm on these two kinds of images and obtain the saliency results, which suggest the RGB image has more information to estimate robust saliency map than grayscale one.

For grayscale-thermal or RGB-T tracking, there are many works. For example, Liu \emph{et al.}~\cite{Sun2012Fusion} performed joint sparse representation calculation on both grayscale and thermal modalities and performed online tracking in Bayesian filtering framework. Li \emph{et al.}~\cite{Li2017Grayscale} utilized the multitask Laplacian sparse representation and integrated modal reliabilities into the model to achieve effective fusion. In~\cite{li2017weighted}, they proposed a patch-based graph model to learn object feature presentation for RGB-T tracking, where the graph is optimized via weighted sparse representations that utilize multi-modality information adaptively. Li \emph{et al.}~\cite{Li18eccv} provided a graph-based cross-modal ranking model for RGB-T tracking, in which the soft cross-modality consistency between modalities and the optimal query learning are introduced to improve the robustness.

\section{Collaborative Graph Learning Algorithm}
In this section, we will introduce the collaborative graph learning algorithm, and describe the details of RGB-T image saliency detection in the next section.

\subsection{Problem Formulation}
Given a pair of RGB and thermal images, we employ SLIC algorithm~\cite{Achanta2012SLIC} to generate $n$ non-overlapping superpixels, where the thermal image is regarded as one of image channels to guarantee segmentation consistency in different modalities. These superpixels are taken as nodes to construct a graph $G=(X,E)$, where $X$ is a node set and $E$ is a set of undirected edges, and extract features from these superpixels, denoting as $X=\{{\bf x}_1,...,{\bf x}_n\}\in\mathbb{R}^{d\times n}$.

If the nodes $i$ and $j$ are spatially adjacent with 8-neighbors, we connect these two nodes and assign an edge weight to it as:
\begin{equation}
\centering
\label{eq::multi-modal weight}
a_{ij}=e^{-\sigma||{\bf x}_i - {\bf x}_j||}.
\end{equation}
where ${a}_{ij}\in [0,1]$, which represents the similarity between the two nodes, ${\sigma}$ is a parameter that controls the strength of the weights. However, the graph of the fixed structure only includes local information and ignores their intrinsic relations, as demonstrated in many vision tasks, such as subspace segmentation~\cite{Guo2015Robust} and visual tracking~\cite{Li18pami}. Therefore, using this kind of graphs might limit the performance of image saliency detection. In this paper, we aim to learn a more meaningful graph which could reflect the ``real'' relationship of graph nodes instead of only spatially adjacent relations as follows:
\begin{equation}
\label{eq::learned W}
\begin{aligned}
&\min_{{\bf W}}\sum_{i,j=1}^{n}{a_{ij}}||{\bf w}_i-{\bf w}_j||^2+\mu\sum_{i=1}^{n}||{\bf w}_i-{\bf i}_i||^2.\\
\end{aligned}
\end{equation}
where ${\bf W}=[{\bf w}_1,{\bf w}_2,...,{\bf w}_n]$ is the learnt affinity matrix based on the structure-fixed graph, in which ${\bf w}_i=[w_{i1},w_{i2},...,w_{in}]^T$ is a column vector that indicates similarities between the node $i$ and other nodes, that is, ${\bf w}_{ij}\in [0,1]$ measures the possibility of $j$ being the true neighbor of $i$. ${\bf i}_i$ is the $i$-th column of an unit matrix $\bf I$, it denoting that the node $i$ is similar to itself at first. The first term indicates that two nodes have analogous similarity                                                                                                                                                                                                                                                                                                                                                                                                                                                                                                                                                                           relationships with all other nodes, i.e., nearby nodes (large $a_{ij}$) should have similar neighbors (small distance between ${\bf w}_i$ and ${\bf w}_j$). The second term is a fitting term, which emphasizes that no matter how we update the indicator ${\bf w}_i$ for node ${\bf x}_i$, it shall still enforce itself as its neighbor as much as possible. The parameter ${\mu}$ controls the balance of two constraints. As demonstrated in~\cite{Bai2016Smooth}, ~\eqref{eq::learned W} is a robust algorithm to select neighbors of graph nodes in an unsupervised way. That makes nearby nodes on the underlying manifold are guaranteed to produce similar neighbors as much as possible. Motivated by this fact, we want to migrate this mechanism to our RGB-T saliency detection to obtain a ``more meaningful'' graph. Thus from~\eqref{eq::learned W}, we could learn a better graph affinity, which is very important in the computation of saliency values~\cite{Yang2013Saliency}. In other words, we determine a good saliency through learning a good graph affinity using~\eqref{eq::learned W}.

\begin{figure}
\includegraphics[width=0.9\columnwidth]{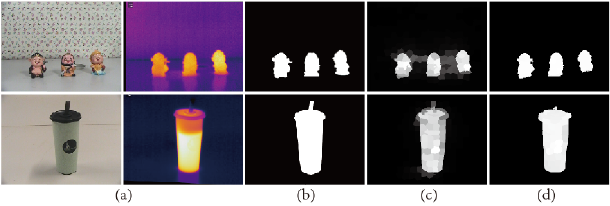}
\centering
\caption{Comparisons of traditional and learnt affinity matrices.
(a) Input RGB-T image pairs. (b) Ground truth. (c) Results produced by the MTMR~\cite{Li2017A}. To be fair, we utilize multiple features (including handcraft color features and deep features) and concatenate them into a single feature vector. (d) Results generated by learnt affinity matrices through our collaborative graph learning.}
\label{fig::graph}
\end{figure}

Note that features of deeper layers contain richer semantic information to localize salient regions while features of shallower layers have much finer structures to retain clearly object boundaries. Therefore, we collaboratively utilize hierarchical deep features and color features from RGB and thermal modalities to learn a more informative and powerful graph. Specifically, we extract CIE-LAB color features and multi-layer deep features from the pre-trained FCN-32S network~\cite{Long2015Fully} for each superpixel and then compute the graph affinities, denoting as $a_{ij}^{(m,k)}$, where $m\in\{1,2,...,M\}$ and $k\in\{1,2,...,K\}$ denote the indexes of modalities and features, respectively. $M$ and $K$ are the number of modalities and features, respectively. Fig.~\ref{fig::motivation} shows the details of a special case, i.e., $M=2,K=3$. Then, we adopt all graph affinity matrices to collaboratively infer a full affinity matrix as follows:
\begin{equation}
\label{eq::learned W1}
\begin{aligned}
&\min_{{\bf W},{\bm \alpha},{\bm \beta}} \sum_{m=1}^{M}{\alpha}^{(m)^{\gamma_1}}\sum_{k=1}^{K}{\beta}^{(m,k)^{\gamma_2}}\sum_{i,j=1}^{n}{a_{ij}^{(m,k)}}||{\bf w}_i-{\bf w}_j||^2\\
&+\mu\sum_{i=1}^{n}||{\bf w}_i-{\bf i}_i||^2.\\
&s.t.\sum_{m=1}^{M}{\alpha}^{(m)}=1, 0\leq{\alpha}^{(m)}\leq1,\\
&\forall m , \sum_{k=1}^{K}{\beta}^{(m,k)}=1, 0\leq{\beta}^{(m,k)}\leq1.\\
\end{aligned}
\end{equation}
where $\bm\alpha\in\mathbb{R}^M$ is a weight vector whose elements represent imaging qualities (\emph{i.e.}, the reliability degree for detecting saliency) of different modalities, and $\bm\beta\in\mathbb{R}^K$ is also a weight vector whose elements indicate feature reliabilities of different layers. Here, $\bm\alpha$ and $\bm\beta$ are used to achieve adaptive fusion of different modalities and features for handling malfunction of some modalities or layer features. The parameter $\gamma_1$ controls the weight distribution across modalities, and the parameter $\gamma_2$ controls the weight distribution across features. From~\eqref{eq::learned W1} we can see that, for each layer of each modality, we construct a structure-fixed graph ($a_{ij}^{(m,k)}$) to represent the relations among superpixels, where spatially adjacent nodes are connected with an edge whose weight is determined by the appearance compatibility of two nodes. For all layers and modalities, we have multiple structure-fixed graphs, and collaboratively employ them to learn an adaptive graph (${\bf W}$), where each structure-fixed graph is assigned with a quality weight to achieve adaptive fusion. Therefore, the collaboration in our model means the adaptive fusion of multiple graphs constructed from different layers and modalities, and its effectiveness is demonstrated in Fig.~\ref{fig::graph}.

In general, the learnt affinity matrix ${\bf W}$ could be used to compute the saliency values via semi-supervised methods, such as manifold ranking~\cite{Yang2013Saliency} and absorbing Markov chain~\cite{Jiang2013Saliency}. It is worth mentioning that there are some methods taking global cues into account to learn an adaptive graph~\cite{Zhu2018Saliency,ZhangAJLL18} for saliency detection, but they perform graph learning first then detect saliency regions based on the computed graph. Different from these works, we propose a one-stage method to integrate the computation of saliency values into the process of graph learning. In particular, we treat saliency values of superpixels as node weights, and jointly learn the graph structure, graph affinity and graph node values (i.e., saliency measure) in a unified optimization framework. Therefore, the final model is proposed as follows:
\begin{equation}
\label{eq::graph learning}
\begin{aligned}
&\min_{{{\bf s}},{\bf W},{\bm \alpha},{\bm \beta}}\frac{1}{2}{\theta}\sum_{i,j=1}^{n}{{\bf w}_{ij}}||{{\bf s}_i-{\bf s}_j}||^2+\lambda||{\bf s}-{\bf y}||^2+\mu\sum_{i=1}^{n}||{\bf w}_i-{\bf i}_i||^2\\
&+\sum_{m=1}^{M}{\alpha}^{(m)^{\gamma_1}}\sum_{k=1}^{K}{\beta}^{(m,k)^{\gamma_2}}\sum_{i,j=1}^{n}{a_{ij}^{(m,k)}}||{\bf w}_i-{\bf w}_j||^2,\\
&s.t.\sum_{m=1}^{M}{\alpha}^{(m)}=1, 0\leq{\alpha}^{(m)}\leq1,\\
&\forall m , \sum_{k=1}^{K}{\beta}^{(m,k)}=1, 0\leq{\beta}^{(m,k)}\leq1.\\
\end{aligned}
\end{equation}
where $\theta$ and $\lambda$ are balanced parameters, ${\bf s}_i$ denotes the weight (i.e., saliency value) of the $i$-th node, and ${\bf y}$ is an indication vector representing the initial graph nodes~\cite{Yang2013Saliency}. Note that our work is a manifold ranking based saliency detection algorithm~\cite{Yang2013Saliency}, i.e., all superpixels are ranked based on the similarity (i.e., graph affinity) to background and foreground queries. The objective is to make saliency values of nodes closer to foreground queries and far away from background queries and these initial queries are specified in Section~\ref{section::ranking}. Therefore, optimizing the model in~\eqref{eq::graph learning} could improve the quality of saliency computation.

\subsection{Model Optimization}
For the sake of notation convenience, the objective function in~\eqref{eq::graph learning} can be rewritten in matrix form :
\begin{equation}
\label{eq::optimal model}
\begin{aligned}
&\min_{{{\bf s}},{\bf W}, {\bm \alpha},{\bm \beta}}\frac{1}{2}{\theta}\sum_{i,j=1}^{n}{{\bf w}_{ij}}||{{\bf s}_i-{\bf s}_j}||^2+\lambda||{\bf s}-{\bf y}||_F^2\\
&+\sum_{m=1}^{M}{{\alpha}^{(m)^{\gamma_1}}}{{\beta}^{(m,k)^{\gamma_2}}}Tr({\bf W}^T{{\bf L}^{(m,k)}}{\bf W})+\mu||{\bf W}-{\bf I}||_F^2.\\
&s.t.\sum_{m=1}^{M}{\alpha}^{(m)}=1, 0\leq{\alpha}^{(m)}\leq1,\\
&\forall m , \sum_{k=1}^{K}{\beta}^{(m,k)}=1, 0\leq{\beta}^{(m,k)}\leq1.\\
\end{aligned}
\end{equation}
where ${\bf L}^{(m,k)}={\bf D}^{(m,k)}-{\bf A}^{(m,k)}$ is the graph Laplacian matrix, and ${\bf D}^{(m,k)}$ and ${\bf A}^{(m,k)}$ are the degree matrix and the graph affinity matrix respectively, calculated by the $k$-th feature at the $m$-th modality. Then we iteratively solve the optimization problem by decomposing it into four sub-problems:

{\flushleft {\bf Solving} ${\bf W}$:}
\\When fixing other variables, we reformulate~\eqref{eq::optimal model} as follows:
\begin{equation}
\label{eq::solve W}
\begin{aligned}
&\min_{\bf W}\sum_{m=1}^{M}{{\alpha}^{(m)^{\gamma1}}}\sum_{k=1}^{K}{{\beta}^{(m,k)^{\gamma2}}}Tr({\bf W}^T{{\bf L}^{(m,k)}}{\bf W})+\mu||{\bf W}-{\bf I}||_F^2\\
&+{\frac{1}{2}{\theta}\sum_{i,j=1}^{n}{{\bf w}_{ij}}||{{\bf s}_i-{\bf s}_j}||^2}.
\end{aligned}
\end{equation}
To compute {\bf W}, the objective function in~\eqref{eq::solve W} can be rewritten in the matrix form:
\begin{equation}
\label{eq::solve WW}
\begin{aligned}
&\min_{\bf W}\sum_{m=1}^{M}{{\alpha}^{(m)^{\gamma1}}}\sum_{k=1}^{K}{{\beta}^{(m,k)^{\gamma2}}}Tr({\bf W}^T{{\bf L}^{(m,k)}}{\bf W})+\mu||{\bf W}-{\bf I}||_F^2\\
&+{\frac{1}{2}{\theta}{\bf W}{\circ}{\bf S}}.\\
&\ \Rightarrow{\bf W}=(\sum_{m=1}^{M}{{\alpha}^{(m)^{\gamma1}}}\sum_{k=1}^{K}{{\beta}^{(m,k)^{\gamma2}}}{\bf L}^{(m,k)}+{\mu}{\bf I})^{-1}{(\mu{\bf I}-\frac{1}{4}{\theta}{\bf S})},
\end{aligned}
\end{equation}
where $\circ$ denotes the element-wise product operation, and ${\bf S}$ is a matrix whose $ij$-th element is $({{\bf s}_i-{\bf s}_j})^2$.

{\flushleft {\bf Solving} ${\bm\beta}$:}
\\By fixing ${\bf W}$, the ${\bm\beta}$-subproblem can be formulated as:
\begin{equation}
\label{eq::solve beta}
\begin{aligned}
&\min_{\bm \beta}\sum_{m=1}^{M}{{\alpha}^{(m)^{\gamma1}}}\sum_{k=1}^{K}{{\beta}^{(m,k)^{\gamma2}}}Tr({\bf W}^T{{\bf L}^{(m,k)}}{\bf W}).\\
&\ \Rightarrow{{\beta}^{(m,k)}}=\frac{(Tr({\bf W}^T{{\bf L}^{(m,k)}}{\bf W}))^{\frac{1}{1-{\gamma2}}}}{\sum_{k=1}^{K}(Tr({\bf W}^T{{\bf L}^{(m,k)}}{\bf W}))^{\frac{1}{1-{\gamma2}}}},
\end{aligned}
\end{equation}

From~\eqref{eq::solve beta}, we can obtain the importance weights of multi-layer features at each modality. ${{\beta}^{(m,k)}}$ is initialized to ${\frac{1}{K}}$ and adaptively updated by~\eqref{eq::solve beta}.

\begin{algorithm}[t]\small
\caption{Optimization Procedure}\label{alg::algorithm}
\begin{algorithmic}[1]
\REQUIRE
Multi-level Laplacian matrices $\{{\bf L}^{(m,k)}\}$, indication vectors $\bf y$, and the parameters $\theta$, $\mu$, $\lambda_1$ and $\gamma_1$,$\gamma_2$;\\
Set $\varepsilon=10^{-4}$ and $maxIter=50$,\\
Initial ${\alpha}^{(m)}=1/M,{\beta}^{(m,k)}=1/K$.
\ENSURE
${\bf s}$, ${\bm \alpha}$, ${\bm \beta}$ and ${\bf W}$.
\FOR{$t=1:maxIter$}
\STATE Update ${\bf W}$ by~\eqref{eq::solve WW};
\STATE Update ${{\beta}^{(m,k)}}$ by~\eqref{eq::solve beta};
\STATE Update ${\alpha}^{(m)}$ by~\eqref{eq::solve alpha};
\STATE Update ${\bf s}$ by~\eqref{eq::solve s};
\IF {Check the convergence condition: the maximum element changes of all variables are lower than $\varepsilon$ or the iteration number reaches $maxIter$}
\STATE Terminate the loop.
\ENDIF
\ENDFOR
\end{algorithmic}
\end{algorithm}

{\flushleft {\bf Solving} ${\bm\alpha}$:}
\\With other variables in~\eqref{eq::optimal model} are fixed, the ${\bm\alpha}$-subproblem can
be written as:
\begin{equation}
\label{eq::solve alpha}
\begin{aligned}
&\min_{\bm\alpha}\sum_{m=1}^{M}{{\alpha}^{(m)^{\gamma1}}}\sum_{k=1}^{K}{{\beta}^{(m,k)^{\gamma2}}}Tr({\bf W}^T{{\bf L}^{(m,k)}}{\bf W}).\\
&\ \Rightarrow{{\alpha}^{(m)}}=\frac{(\sum_{k=1}^{K}{{\beta}^{(m,k)^{\gamma2}}}Tr({\bf W}^T{{\bf L}^{(m,k)}}{\bf W}))^{\frac{1}{1-{\gamma1}}}}{\sum_{m=1}^{M}(\sum_{k=1}^{K}{{\beta}^{(m,k)^{\gamma2}}}Tr({\bf W}^T{{\bf L}^{(m,k)}}{\bf W}))^{\frac{1}{1-{\gamma1}}}},
\end{aligned}
\end{equation}

From~\eqref{eq::solve alpha}, we can obtain the importance weights of each modality. ${{\alpha}^{(m)}}$ is initialized to ${\frac{1}{M}}$ and adaptively updated by~\eqref{eq::solve alpha}.

{\flushleft {\bf Solving} ${\bf s}$:}
\\When other variables in~\eqref{eq::optimal model} are fixed, the ${\bf s}$-subproblem
can be formulated as:
\begin{equation}
\begin{aligned}
\label{eq::solve s1}
&\min_{\bf s}{\theta}\sum_{i,j=1}^{n}{{\bf w}_{ij}}||{{\bf s}_i-{\bf s}_j}||^2+\lambda||{\bf s}-{\bf y}||_F^2.
\end{aligned}
\end{equation}
To compute ${\bf s}$, the objective function in~\eqref{eq::solve s1} can be rewritten in the matrix form :
\begin{equation}
\label{eq::solve s}
\begin{aligned}
&\min_{\bf s}{\theta}{\bf s}^T({\bf D}-{\bf W}){\bf s}+\lambda||{\bf s}-{\bf y}||_F^2.\\
&\ \Rightarrow{\bf s}=({\lambda_1}{\bf F}+{\bf I})^{-1}{\bf y},
\end{aligned}
\end{equation}
where ${\lambda_1}={\frac{\theta}{\lambda}}$, ${\bf F}={\bf D}-{\bf W}$ is the Laplacian matrix, ${\bf D}_{ii}={\sum_{j=1}^{n}}{{\bf W}_{ij}}$, where ${\bf W}$ and ${\bf D}$ are the learnt graph affinity matrix and its degree matrix, respectively. We summarize whole optimization procedure in Algorithm~\ref{alg::algorithm}. Since each subproblem of~\eqref{eq::optimal model} is convex, the solution by the proposed algorithm satisfies the Nash equilibrium conditions~\cite{Xu2015A}.

{\flushleft \bf Complexity analysis.} It's worth noting that the complexity of each sub-problem is ${O(n^3)}$, where $n$ is the size of ${\bf W}$. Denoting the number of iterations as $T$, the overall complexity of our optimization algorithm is ${O(Tn^3)}$.

\section{Two-stage RGB-T Saliency Detection}
\label{section::ranking}
In this section, we present the two-stage ranking scheme for bottom-up RGB-T saliency detection using background and foreground queries.

{\flushleft \bf Ranking with Background Queries}.
First, we utilize the boundary superpixels as initial background queries widely used in other works~\cite{Yang2013Saliency,Wang2016GraB} for our approach to highlight the salient superpixels, and select high confident superpixels (low ranking scores in all modalities) belonging to the foreground as the foreground queries. Specifically, we construct four saliency maps using boundary priors and then integrate them for the final map, which is referred as the separation/combination (SC) approach~\cite{Yang2013Saliency}.

Taking the boundary on the top as an example, we regard the top boundary superpixels as the background queries, and the other nodes as the unlabeled superpixels. We run our algorithm to obtain the ranking vector ${\bf s}$ and then normalize it to the range between 0 and 1. It notes that our graph learning and saliency inference in a unified framework. The saliency map ${\bf s}_t$ with the top boundary queries is computed by:
\begin {center}
${\bf s}_t=1-{\bf \hat s}$.
\end {center}

Similarly, we can obtain other ranking vectors with bottom, left, right boundary superpixels, denoting as ${\bf  s}_b$, ${\bf  s}_l$, ${\bf  s}_r$, respectively, and the final saliency map ${\bf s}{_{bq}}$ with background queries is computed as follows:
\begin {center}
${\bf s}{_{bq}}={{\bf s}_t}{\circ}{{\bf  s}_b}{\circ}{{\bf s}_l}{\circ}{{\bf  s}_r}$,
\end {center}
where $\circ$ indicates the element-wise product operation.

{\flushleft \bf Ranking with Foreground Queries}.
Given ${\bf s}{_{bq}}$, we set an adaptive threshold to generate foreground queries, then utilize these for graph learning and infer the saliency map in a unified framework. At last, the final saliency map can be obtained by normalizing the saliency map with foreground queries into the range of 0 and 1.

{\flushleft \bf Difference from Previous Work}.
For the problem of RGB-T image saliency detection, there is only one work on RGB-T saliency detection~\cite{Li2017A}, and it should be noted that our work is significantly different from theirs. In~\cite{Li2017A}, they proposed a multi-task graph-based manifold ranking model that only uses handcrafted features and structure-fixed graphs, and also propose a new RGB-T dataset for saliency detection purpose. While we employ multi-level deep features and structure-fixed graphs to learn a more powerful collaborative graph to better explore intrinsic relations among graph nodes. There are some methods to learn adaptive graphs for saliency detection~\cite{Zhu2018Saliency,ZhangAJLL18}, but they usually perform two steps for saliency computation. Different from these works, we integrate these two steps into a joint process and propose a one-stage method for further boosting their respective performance. In addition, we also contribute a more comprehensive RGB-T dataset for saliency detection purpose, and the advantages over~\cite{Li2017A} are presented in Section~\ref{subsec::advantages}.

\section{VT1000 Dataset}
\label{section::Dataset}
In this work, considering the data deficiency and in order to promote the research for RGB-T saliency detection, we capture 1000 image pairs including visible images and their corresponding thermal maps in diverse scenes, named VT1000 in this paper. In this section, we introduce the new dataset with statistic analysis.

\begin{figure}
\includegraphics[width=0.9\columnwidth]{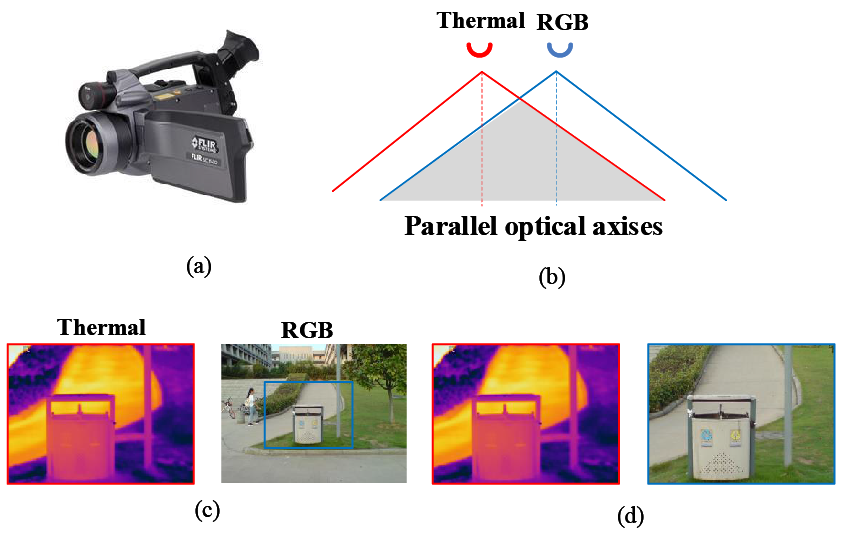}
\caption{The mechanism of our imaging platform for RGB-T image pairs.}
\label{fig::register}
\end{figure}

\subsection{Platform}
Our imaging hardware is FLIR (Forward Looking Infrared) SC620, with a thermal infrared camera and a CCD camera inside, as shown in Fig.~\ref{fig::register} (a). It means that the two cameras have the same imaging parameters except for focus, and their optical axes are aligned as parallel, as shown in Fig.~\ref{fig::register} (b). Then, we make the image alignment manually by enlarging the visible image and crop the part of the visible image and cropping the visible image to totally overlap with the thermal infrared image. Fig.~\ref{fig::register} (c) shows the visible image and thermal infrared image before alignment, where the annotated bounding boxes denote the common horizon. Fig.~\ref{fig::register} (d) shows the highly aligned RGB-T image pairs.

\begin{table}[t]\footnotesize
\centering
\caption{List of the annotated challenges of our dataset.}\label{tb::challenges}
\begin{tabular}{|c|p{180pt}|}
\hline
\textbf{Challenges} & \textbf{Description}\\
\hline
\hline
BSO & Big Salient Object - the ratio of ground truth salient objects over image is more than 0.26.\\
SSO & Small Salient Object - the ratio of ground truth salient objects over image is less than 0.05.\\
LI & Low Illumination - the environmental illumination is low.\\
MSO & Multiple Salient Objects - the number of the salient objects in the image is greater than 1.\\
CB & Center Bias - the center of the salient object is further away from the center of the image. \\
CIB & Cross Image Boundary - the salient objects cross the image boundaries. \\
SA & Similar Appearance - the salient object has similar color or shape to the background surroundings. \\
TC & Thermal Crossover - the salient object has similar temperature with other objects or background surroundings. \\
IC & Image Clutter - the background information which includes the target object is clutter. \\
OF & Out of Focus - the object in image is out-of-focus, the entire image is fuzzy. \\
\hline
\end{tabular}
\end{table}

\begin{table*}[t]\footnotesize
\setlength{\belowcaptionskip}{0.2cm}
\caption{Distribution of visual attributes within the VT1000 dataset, showing the number of coincident attributes across all RGB-T image pairs.}
\centering
\begin{tabular}{ c | c  c  c  c  c  c  c  c  c  c}
	\hline
 & BSO& CB & CIB& IC& LI & MSO & OF & SA & SSO & TC\\\hline
 BSO &\textbf{145}&2&18&18&5&7&25&6&0&48\\
 CB &2&\textbf{169}&34&38&6&4&18&20&49&63\\
 CIB &18&34&\textbf{125}&43&3&7&4&15&12&28\\
 IC &18&38&43&\textbf{162}&1&13&5&4&27&86\\
 LI &5&6&3&1&\textbf{56}&0&10&2&6&14\\
 MSO&7&4&7&13&0&\textbf{87}&3&10&2&14\\
 OF&25&18&4&5&10&3&\textbf{122}&17&21&34\\
 SA &6&20&15&4&2&10&17&\textbf{142}&21&11\\
 SSO&0&49&12&27&6&2&21&21&\textbf{183}&59\\
 TC&48&63&28&86&14&14&34&11&59&\textbf{282}\\
\hline
\end{tabular}
\label{tb::Attribute distribution}
\end{table*}

\subsection{Annotation}
For better evaluate the RGB-T saliency detection, we capture 2000 natural RGB-T image pairs approximately, and we manually selected 1500 image pairs.
Then, for each selected image, six participants are requested to choose their first glance at the most salient object. Since different people have different views on what a salient object is in the same image, we get rid of those images with low labelling consistency and select top 1000 image pairs.
Finally, four participants use Adobe Photoshop to crop the RGB images that are totally overlapped with thermal images, and then segment the salient object manually from each image to obtain pixel-level ground truth masks.

\subsection{Statistics}
The image pairs in our dataset are recorded under different illuminations, object categories, sizes and numbers, \emph{etc}. we annotate 10 challenges in our dataset to facilitate the challenge-sensitive performance evaluation of different algorithms. They are: big salient object (BSO), small salient object (SSO), multiple salient object (MSO), low illumination (LI), center
bias (CB), cross image boundary (CIB), similar appearance(SA), thermal crossover (TC), image clutter (IC), and out of focus (OF). In particular, Table~\ref{tb::challenges} shows the details. We also present the attribute distribution over the VT1000 dataset in Table~\ref{tb::Attribute distribution}.

\begin{figure}[t]
\includegraphics[width=0.6\columnwidth]{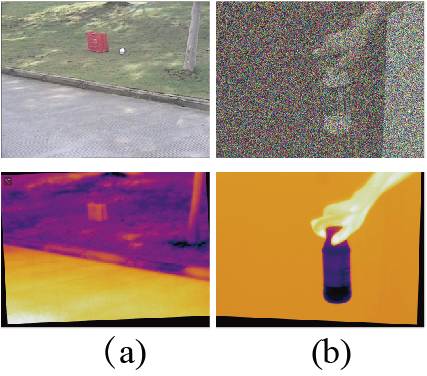}
\centering
\caption{Two examples RGB-T data of the VT821 dataset, (a) and (b) indicate aligned RGB and thermal image.}
\label{fig::MTMRflaw}
\end{figure}

\begin{figure*}[t]
\includegraphics[width=\textwidth]{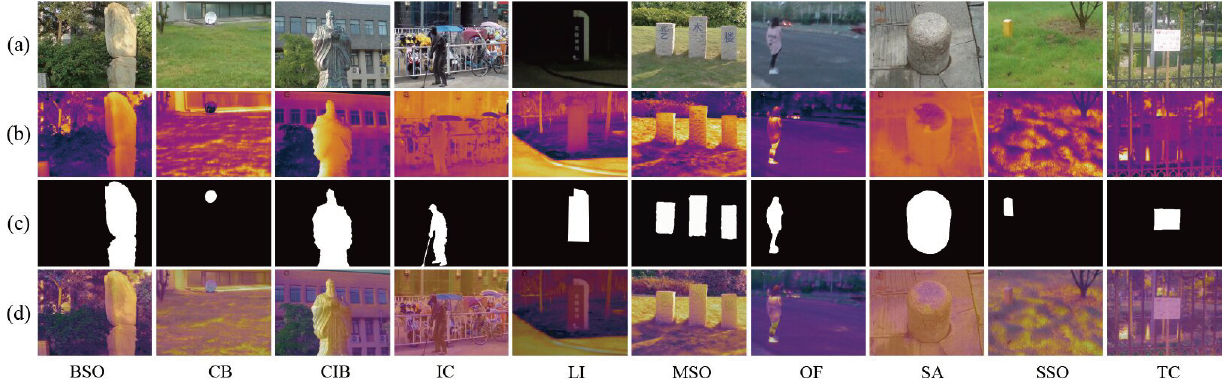}
\centering
\caption{Sample image pairs with annotated ground truths and challenges from our RGB-T dataset. (a) and (b) indicate RGB and thermal image. (c) is corresponding
ground truth of RGB-T image pairs. (d) is the fused result based on RGB-T image pairs.}
\label{fig::subset}
\end{figure*}

\subsection{Advantages over existing Datasets}
\label{subsec::advantages}%% 应该先介绍下其他的数据集的一些内容 再说我们的数据集

There are many datasets for salient object detection, existing datasets are limited in their coarse annotation for salient objects and the number of images. For improving the quality of datasets, recently, researchers start to construct the datasets with objects in relatively complex and cluttered backgrounds,  such as DUT-OMRON~\cite{Yang2013Saliency}, ECSSD~\cite{Yan2013Hierarchical}, Judd-A~\cite{Borji2012Salient}, and PASCAL-S~\cite{Li2014The}. Compared with their predecessors, these datasets have been improved in terms of annotation quality and image number. Since RGB spectrum is sensitive to light and depend too much on “good” lighting and environmental conditions, it could be easily affected by bad environments, like smog, rain and fog. Meanwhile, the thermal infrared data are more effective in capturing objects than visible spectrum cameras under poor lighting conditions and bad weathers. However, its weakness is revealed when the thermal crossover occurs.
Taking into account the aforementioned limitations of existing datasets, it is necessary to construct a unified RGB-T dataset that enables salient object detection in more complicated conditions. Compared to the only existing RGB-T dataset~\cite{Li2017A} (called VT821 in this paper), our VT1000 dataset has the following advantages.

\begin{itemize}

\item Our dataset includes more than 400 kinds of common objects collected in 10 types of scenes under different illumination conditions. The indoor scenes include offices, apartments, supermarkets, restaurant, library, \emph{etc}. While outdoor locations contain parks, campuses, streets, buildings, lakes, \emph{etc}. Most images contain a single salient object, while the others include multiple objects. Compared to VT821, our dataset is larger and challenging.

\item Since the imaging parameters of RGB and thermal cameras in our platform are basically the same and their optical axes are parallel, its images can be well captured by static or moving cameras. The transformation between two modal images are only translation and scale, which also makes the images with different modalities can be highly aligned, and without any noise in the boundary. Fig.~\ref{fig::MTMRflaw} shows two RGB-T image pairs in the VT821 dataset, which contain blank boundaries in thermal modality caused by their alignment method.

\item It's worth noting that we collect the VT1000 in the summer, which causes the high surface temperature of most objects in the scenes. Many thermal images appear thermal crossover, which increases challenges of our VT1000 dataset. We take these challenges into consideration and divide these with 10 attributes for occlusion-sensitive evaluation of different algorithms as same as VT821. As shown in the Fig.~\ref{fig::subset}, we present some sample image pairs with ground truth and attribute annotations. For clarity, we also present some fused RGB-T images to indicate highly aligned results of different modalities.

\end{itemize}

%%1000
\begin{table*}[t]\footnotesize
  \centering
  \caption{Average precision, recall, and F-measure of our method against different kinds of baseline methods on the VT1000 Dataset, where the best, the second and the
third best results are in \textcolor{red}{red}, \textcolor{green}{green} and \textcolor{blue}{blue} colors, respectively.}
  \label{tab:freq2}
  \begin{tabular}{|l||c|c|c||c|c|c||c|c|c||c|c|}
\hline
\multirow{2}{*}{\textbf{Algorithm}} & \multicolumn{3}{c||}{\textbf{RGB}} & \multicolumn{3}{c||}{\textbf{Thermal}} &
\multicolumn{3}{c||}{\textbf{RGB-T}} \\
\cline{2-10}
 & $P$ & $R$ & $F$  & $P$ & $R$ & $F$ & $P$ & $R$ & $F$\\
\hline
  MR~\cite{Yang2013Saliency} &0.766&0.588&0.635&{\textcolor{green}{0.706}}&0.555&0.586&0.759&0.600&0.635\\
\hline
  RBD~\cite{Zhu2014Saliency}   &0.717&{\textcolor{green}{0.680}}&0.628&0.649&{\textcolor{red}{0.677}}&0.576&0.718&{\textcolor{red}{0.745}}&0.650\\
\hline
  CA~\cite{Qin2015Saliency}&0.718&0.644&0.621&0.676&0.598&0.581&0.701&0.636&0.610\\
\hline
  RRWR~\cite{Li2015Robust}   &0.766&0.594&0.637&{\textcolor{blue}{0.703}}&0.557&{\textcolor{green}{0.596}}&0.584&0.592&0.616\\
\hline
 MILPS~\cite{Huang2017Salient}	&{\textcolor{blue}{0.769}}&0.664&{\textcolor{blue}{0.663}}&{\textcolor{red}{0.714}}&{\textcolor{green}{0.608}}&{\textcolor{red}{0.610}}&0.762&{\textcolor{blue}{0.686}}&0.661\\
\hline
FCNN~\cite{Li2016DeepSaliency}   &{\textcolor{green}{0.771}}&{\textcolor{red}{0.746}}&{\textcolor{green}{0.689}}&0.688&{\textcolor{blue}{0.635}}&{\textcolor{blue}{0.590}}&0.750&{\textcolor{green}{0.739}}&{\textcolor{blue}{0.671}}\\		
\hline
DSS~\cite{HouCvpr2017Dss}   &{\textcolor{red}{0.877}}&{\textcolor{blue}{0.676}}&{\textcolor{red}{0.721}}&0.660&0.357&0.416&{\textcolor{green}{0.808}}&0.594&0.634\\			
\hline
 MTMR~\cite{Li2017A}  &-&-&-&-&-&-&{\textcolor{blue}{0.792}}&0.627&{\textcolor{green}{0.673}}\\
\hline
Ours         &-&-&-&-&-&-&{\textcolor{red}{0.853}}&0.649&{\textcolor{red}{0.727}}\\
\hline
\end{tabular}
\end{table*}

\begin{table*}\footnotesize
\setlength{\belowcaptionskip}{0.2cm}
\caption{F-measure based on different attributes of the proposed approach with 8 methods on the VT1000 dataset, where the best, the second and the
third best results are in \textcolor{red}{red}, \textcolor{green}{green} and \textcolor{blue}{blue} colors, respectively.}
\label{tb::A2}
\centering
\begin{tabular}{ c | c  c  c  c  c  c  c  c  c}
	\hline
 & MR& RBD& CA & RRWR & MILPS & FCNN & DSS & MTMR & ours\\
 \hline
 BSO &0.750&{\textcolor{red}{0.813}}&{\textcolor{green}{0.796}}&0.735&0.717&{\textcolor{blue}{0.794}}& 0.613& 0.741& 0.771\\
 CB &0.468&0.488&0.413&0.458&0.499&{\textcolor{blue}{0.551}}&{\textcolor{green}{0.609}}&0.541&{\textcolor{red}{0.627}}\\
 CIB &0.572&0.606&0.552&0.534&{\textcolor{blue}{0.644}}&{\textcolor{green}{0.675}}&0.632&0.565&{\textcolor{red}{0.693}}\\
 IC &0.506&0.460&0.486&0.458&0.528&{\textcolor{blue}{0.591}}&{\textcolor{green}{0.594}}&0.520&{\textcolor{red}{0.627}}\\
 LI &0.626&0.646&{\textcolor{green}{0.671}}&0.647&0.615&{\textcolor{red}{0.680}}&0.423 &0.647&{\textcolor{blue}{0.648}}\\
 MSO&0.690&0.724&0.716&0.681&0.732&{\textcolor{green}{0.754}}&0.713&{\textcolor{blue}{0.739}}&{\textcolor{red}{0.773}}\\
 OF&0.580&0.640&0.602&{\textcolor{blue}{0.645}}&0.609&0.632&{\textcolor{red}{0.713}}&0.627&{\textcolor{green}{0.669}}\\
 SA &0.621&{\textcolor{blue}{0.705}}&0.664&0.686&0.700&{\textcolor{green}{0.723}}&0.437&0.703 &{\textcolor{red}{0.753}}\\
 SSO&0.444&0.456&0.312&0.415&0.479&0.425&{\textcolor{green}{0.603}}&{\textcolor{blue}{0.556}}&{\textcolor{red}{0.681}}\\
 TC&0.584&0.543&0.513&0.508&0.577&{\textcolor{green}{0.605}}&0.573 &{\textcolor{blue}{0.594}}&{\textcolor{red}{0.670}}\\
 \hline
 Entire   &0.635&0.650&0.610&0.616&0.661&{\textcolor{blue}{0.671}}&0.634&{\textcolor{green}{0.673}}&{\textcolor{red}{0.727}} \\
\hline
\end{tabular}
\end{table*}

%%%VT821
\begin{table*}\footnotesize
  \centering
  \caption{Average precision, recall, and F-measure of our method against different kinds of baseline methods on the VT821 dataset, where the best, the second and the
third best results are in \textcolor{red}{red}, \textcolor{green}{green} and \textcolor{blue}{blue} colors, respectively.}
  \label{tab:freq1}
  \begin{tabular}{|l||c|c|c||c|c|c||c|c|c||c|c|}
\hline
\multirow{2}{*}{\textbf{Algorithm}} & \multicolumn{3}{c||}{\textbf{RGB}} & \multicolumn{3}{c||}{\textbf{Thermal}} &
\multicolumn{3}{c||}{\textbf{RGB-T}} \\
\cline{2-10}
 & $P$ & $R$ & $F$  & $P$ & $R$ & $F$ & $P$ & $R$ & $F$\\
\hline
  MR~\cite{Yang2013Saliency}  &{\textcolor{green}{0.644}}&0.603&0.587&{\textcolor{red}{0.700}}&0.574&{\textcolor{blue}{0.603}}&{\textcolor{green}{0.733}}&0.653&{\textcolor{blue}{0.666}}\\
\hline
  RBD~\cite{Zhu2014Saliency}   &0.612&{\textcolor{green}{0.738}}&0.603&0.550&{\textcolor{red}{0.784}}&0.556&0.612&{\textcolor{red}{0.841}}&0.622 \\
\hline
 CA~\cite{Qin2015Saliency}    &0.593&0.668&0.569&0.625&0.612&0.577&0.645&0.668&0.609\\
\hline
RRWR~\cite{Li2015Robust}   &{\textcolor{blue}{0.642}}&0.610&0.589&{\textcolor{green}{0.689}}&0.580&0.596&0.695&0.617&0.628\\
\hline
MILPS~\cite{Huang2017Salient}  &0.637&0.691&{\textcolor{blue}{0.612}}&{\textcolor{blue}{0.643}}&{\textcolor{blue}{0.680}}&{\textcolor{green}{0.612}}&0.664&{\textcolor{blue}{0.753}}&0.644\\
\hline
 FCNN~\cite{Li2016DeepSaliency}  &0.636&{\textcolor{red}{0.806}}&{\textcolor{green}{0.642}}&0.627&{\textcolor{green}{0.711}}&{\textcolor{red}{0.615}}&0.647&{\textcolor{green}{0.820}}&0.653\\				\hline
 DSS~\cite{HouCvpr2017Dss}  &{\textcolor{red}{0.740}}&{\textcolor{blue}{0.727}}&{\textcolor{red}{0.693}}&0.462&0.240&0.307&0.710&0.673&0.639\\				\hline
 MTMR~\cite{Li2017A}  &-&-&-&-&-&-&{\textcolor{blue}{0.716}}&0.713&{\textcolor{green}{0.680}}\\
\hline
Ours         &-&-&-&-&-&-&{\textcolor{red}{0.794}}&0.724&{\textcolor{red}{0.744}}\\
\hline
\end{tabular}
\end{table*}

\begin{table*}\footnotesize
\setlength{\belowcaptionskip}{0.2cm}
\caption{F-measure based on different attributes of the proposed approach with 8 methods on the VT821 dataset, where the best, the second and the
third best results are in \textcolor{red}{red}, \textcolor{green}{green} and \textcolor{blue}{blue} colors, respectively.}
\label{tb::A1}
\centering
\begin{tabular}{ c | c  c  c  c  c  c  c  c  c}
	\hline
  & MR & RBD& CA & RRWR & MILPS & FCNN & DSS & MTMR & ours\\\hline
 BSO & 0.797& {\textcolor{red}{0.843}}& {\textcolor{blue}{0.809}}& 0.756 &0.772& 0.766& 0.593 & 0.788& {\textcolor{green}{0.817}} \\
 CB & {\textcolor{blue}{0.731}}& 0.692& 0.695 & 0.712 &0.729& 0.721& 0.665 &{\textcolor{green}{0.750}}& {\textcolor{red}{0.789}} \\
 CIB & 0.634& {\textcolor{blue}{0.692}}& 0.597 & 0.581 &0.641& {\textcolor{red}{0.727}}& 0.645 & 0.628& {\textcolor{green}{0.699}}\\
 IC & 0.591& 0.536& 0.539& 0.548 &0.579&{\textcolor{green}{0.629}}& 0.580& {\textcolor{blue}{0.607}}& {\textcolor{red}{0.689}} \\
 LI & 0.658&0.621& {\textcolor{blue}{0.660}}& 0.651 &0.641 & 0.659 & 0.618 & {\textcolor{green}{0.678}}& {\textcolor{red}{0.723}}\\
 MSO& 0.642 &0.628& 0.613& 0.608 &0.651 & {\textcolor{green}{0.690}}& {\textcolor{blue}{0.670}}& 0.666& {\textcolor{red}{0.737}}\\
 OF & {\textcolor{green}{0.689}}&0.659& 0.638& 0.654 &0.624 & 0.655 & 0.498 &{\textcolor{blue}{0.672}}& {\textcolor{red}{0.722}}\\
 SA & 0.587& 0.552& 0.603& {\textcolor{blue}{0.620}}&0.599 & 0.596 & 0.607 & {\textcolor{green}{0.664}}& {\textcolor{red}{0.699}}\\
 SSO& 0.328& 0.341& 0.238 & 0.275 & 0.247 & 0.259 & {\textcolor{green}{0.444}}& {\textcolor{blue}{0.413}}& {\textcolor{red}{0.513}}\\
 TC& 0.608& 0.592 & 0.561& 0.567 &0.617 &{\textcolor{green}{0.655}}& {\textcolor{blue}{0.638}}& 0.628&{\textcolor{red}{0.713}}\\\hline
 Entire & {\textcolor{blue}{0.666}}& 0.622 & 0.609& 0.628 &0.644 &0.653& 0.639 &{\textcolor{green}{0.680}}&{\textcolor{red}{0.744}}\\
\hline
\end{tabular}
\end{table*}

\section{Experiments}
We evaluate the proposed approach on the public VT821 dataset~\cite{Li2017A} and the newly created VT1000 dataset.
In this section, we will present the experimental results of the proposed approach on the two RGB-T datasets, and then compare with other state-of-the-art methods. At last, our approach components are analyzed in detail.
\begin{figure}[ht]
\centering
\includegraphics[width=0.4\textwidth]{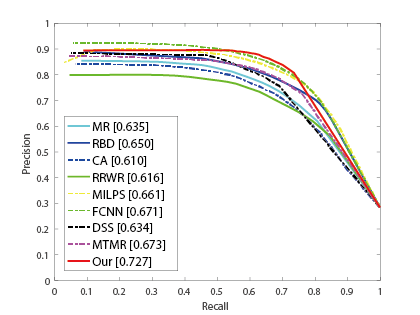}
\caption{P-R curves of the proposed approach and other methods with RGB-T inputs on the VT1000 dataset. The representative score of F-measure is presented in the legend.}
\label{fig::pr2}
\end{figure}

\begin{figure}[ht]
\centering
\includegraphics[width=0.4\textwidth]{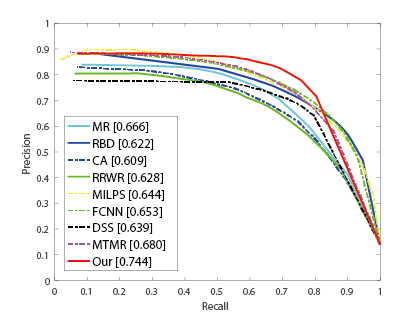}
\caption{P-R curves of the proposed approach and other methods with RGB-T inputs on the VT821 dataset. The representative score of F-measure is presented in the legend.}
\label{fig::pr1}
\end{figure}

\subsection{Experimental Setup}
{\flushleft \bf Evaluation criteria}.
In our work, we utilize two evaluation metrics to evaluate the performance of our method with other state-of-the-art salient object detection methods, including Precision-Recall (PR) curves, F-measure score. The PR curves are obtained by binarizing the saliency map using thresholds in the range of 0 and 255, where the precision (P) is the ratio of retrieved salient pixels to all pixels retrieved, and the recall (R) is the ratio of retrieved salient pixels to all salient pixels in the image. Also, we utilize the F-measure (F) to evaluate the quality of a saliency map, which is formulated by a weighted combination of Precision and Recall:
\begin{equation}
F_{\beta}=\frac{(1+\beta^2)\times Precision\times Recall}{\beta^2\times Precision + Recall},
\end{equation}
where we set the $\beta^2 = 0.3$ to emphasize the precision as suggested in~\cite{achanta2009frequency}.

{\flushleft \bf Baseline methods}.
For comprehensively validating the effectiveness of our approach, we qualitatively and quantitatively compare the proposed approach with 8 state-of-the-art approaches, including MR~\cite{Yang2013Saliency}, RBD~\cite{Zhu2014Saliency}, CA~\cite{Qin2015Saliency}, RRWR~\cite{Li2015Robust}, MILPS~\cite{Huang2017Salient}, FCNN~\cite{Li2016DeepSaliency}, DSS~\cite{HouCvpr2017Dss}, MTMR~\cite{Li2017A}. It is worth noting that FCNN and DSS are deep learning based methods.
Comparing with above methods with either RGB or thermal infrared input, we can justify the effectiveness of complementary benefits from different modalities of our approach.
In addition, we implement some RGB-T baselines extended from RGB ones for fair comparison. In a specific, we concatenate the features extracted from RGB and thermal images together as the RGB-T feature representations, and then run RGB saliency detection algorithms to obtain RGB-T results.

\begin{figure*}
  \includegraphics[width=0.9\textwidth]{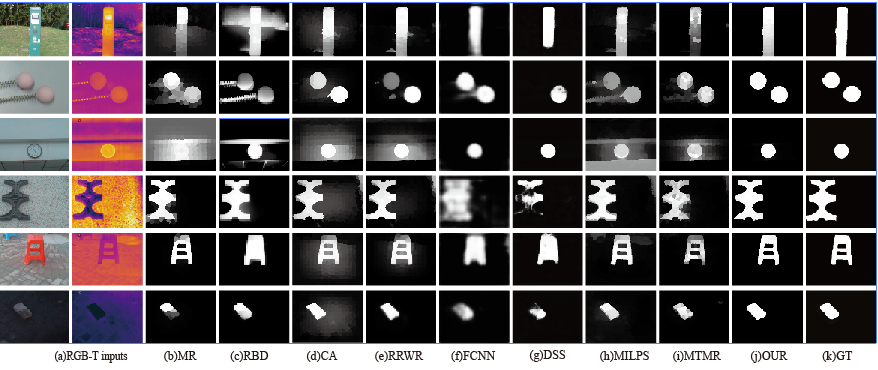}
    \centering
 \caption{(a) Sample results of our approach against other baseline methods, where first two columns are RGB-T inputs. (b-h) show respectively RGB-T saliency detection results generated by the extended RGB-T approaches. (i) shows RGB-T saliency detection results generated by a RGB-T approach. (j) is the results by our proposed approach, (k) is ground truth.}
  \label{fig::visualRGBTfuse}
  \end{figure*}

\begin{figure*}[t]
\includegraphics[width=0.9\textwidth]{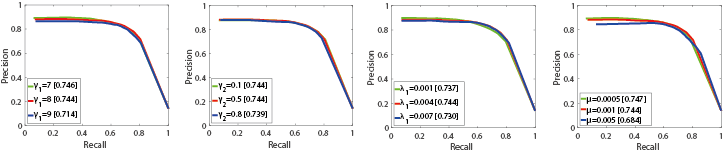}
\centering
\caption{Precision-recall curves on the VT821 dataset by the proposed algorithm with different parameter values. The representative score of F-measure is presented in the legend.}
\label{fig::Params}
\end{figure*}

{\flushleft \bf Parameter settings}.
For fair comparison, we fix all parameters and other settings of our approach in experiments. In graph construction, we fix the number of superpixels to $n=300$. The graph edges are strengthened by the parameter $\sigma$, and we set $\sigma_{RGB}$=20, $\sigma_{T}$=40.

The proposed model involves five parameters, and we set them as: $\{\gamma_1,\gamma_2,\theta,\mu,\lambda_1\}=\{0.5,8,0.0001,0.001,0.004\}$. ${\gamma_1}$ controls the weight distribution across multiple modalities, and the parameter ${\gamma_2 }$ controls the weight distribution of multiple affinity graphs in feature space. Take ${\gamma_2}$ as an example, when ${\gamma_2}\rightarrow$1, only the smoothest affinity graph is computed. When ${\gamma_2}\rightarrow \infty$, equal weights are obtained. The selection of ${\gamma_2}$ mainly depends on the degree of complementary quality among these affinity graphs. Rich complementarity tends to a bigger ${\gamma_2}$, which can make better graph learning of multiple graphs. However, RGB and thermal data are not always optimal, we need to make full use of the complementary information of the two modalities. Therefore, for combining two affinity matrices of two modalities well, we set ${\gamma_1}$ smaller, which can keep the better modality obtaining the higher weight. Note that their variations do not affect the performance much, and we demonstrate their insensitivity in Fig.~\ref{fig::Params}.
To compute the learnt matrix, we use FCN-32S~\cite{Long2015Fully} features that perform well in semantic segmentation, and only adopt the outputs from \emph{Conv1} and \emph{Conv5} layers as feature maps. Since \emph{Conv1} of CNNs encodes low-level detailed features, which can refine the outline of the saliency map, and the \emph{Conv5} carries higher-level semantic features, which keep the object highlight. These two kinds of features have 64 and 512 dimensions, respectively. Note that we utilize the pre-trained FCN-32S network to extract multi-layer feature maps, then resize them (shallow and deep layers) to the size of input image via bilinear interpolation. The feature representation of each superpixel can be computed by averaging features of all pixels within this superpixel.

\subsection{Evaluation on the VT1000 Dataset}
{\flushleft \bf Overall performance.}
We first compare our approach against other methods mentioned above on the aspects of precision (P), recall (R) and F-measure (F), shown in Table~\ref{tab:freq2}. From the quantitative evaluation results, we can observe that the proposed method achieves a good balance of precision and recall, and thus obtains better F-measure. Fig.~\ref{fig::pr2} shows that our method outperforms others with a clear margin. Our method significantly outperforms the latest RGB-T method MTMR~\cite{Li2017A}, achieving 5.4$\%$ gain in F-measure over it. At the same time, it greatly exceeds other extended RGB-T methods. The visual comparison is shown in Fig.~\ref{fig::visualRGBTfuse}, which suggests that our method makes RGB and thermal data effectively fused. We can see that the proposed algorithm highlights the salient regions and has well-defined contours.
{\flushleft \bf Comparison with traditional RGB methods.}
We compare our method with some state-of-the-art traditional RGB saliency detection methods, including MR, RBD, CA, RRWR ,MILPS. Table~\ref{tab:freq2} shows that the F-measure of ours is better than the results with RGB information only, indicating the effectiveness of the introduction of thermal information for image saliency detection. In particular, our method(RGB-T images as input) outperforms MR(RGB images as input) and MILPS(RGB images as input) with 9.2$\%$, 6.4$\%$ in F-measure, respectively. MR is our baseline and the MILPS is the latest traditional RGB method. Therefore, the results in Table~\ref{tab:freq2} verify the effectiveness of our method by collaboratively fusing RGB and thermal information to address challenging scenarios. Comparing with above methods with either RGB or thermal input, we could justify the effectiveness of complementary benefits from different modalities of our approach.

{\flushleft \bf Comparison with RGB-T methods.}
We further compare our method with several traditional extended RGB-T methods (MR, RBD, CA, RRWR, MILPS) and RGB-T method MTMR in Table~\ref{tab:freq2} and Fig.~\ref{fig::pr2}. It is worth mentioning that the RGB-T feature representations in the extended RGB-T method are implemented by concatenating the features extracted respectively from RGB and thermal modalities together. As shown in the comparison with the above six traditional RGB-T methods, our approach obtains higher F-measure scores than other RGB-T methods in Fig.~\ref{fig::pr2}. It is possible not the most optimal to directly concatenate the features of the two modalities. Therefore, it is necessary to consider an adaptive strategy to merge multi-modality features. It is easy to see that our method performs better than the previous graph-based methods such as MR, RRWR and latest RGB-T method MTMR, overcoming them with 9.2$\%$, 11.1$\%$, 5.4$\%$ in F-measure, respectively. These results demonstrate the effectiveness of the proposed approach that employs RGB and thermal information adaptively to learn graph affinity via collaborative graph learning, which is helpful to improving the robustness.

{\flushleft \bf Comparison with deep learning methods.}
We also evaluate with some state-of-the-art deep learning based methods, including FCNN~\cite{Li2016DeepSaliency} and DSS~\cite{HouCvpr2017Dss}. For a fair comparison, we extend the two methods into RGB-T methods by concatenating the features extracted from RGB and thermal modalities together as RGB-T feature representations. Overall, our approach obtains the best performance, as shown in Fig.~\ref{fig::pr2}. In particular, our method outperforms $5.6\%$ over FCNN and 9.3$\%$ over DSS in F-measure. The good results are due to the model jointly learning a collaborative graph of two modalities in a unified optimization framework. Our P-R curve seems slightly lower than FCNN because FCNN achieves higher recalls, it is worth to mention that our method exceeds FCNN on precision and F-measure obviously, as shown in Table~\ref{tab:freq2} , the reason for that is deep learning models are trained on a large amount of RGB images owing to the limited amount of RGB-T images. In addition, our method also has the following advantages over the deep learning based
methods. i) It does not need laborious pre-training or a large training set. ii) It does not need to save a large pre-trained deep model. iii) It is easy to implement as every subproblem of our proposed model has a closed-form solution. iv) It performs favorably against FCNN and DSS in terms of efficiency on a cheaper hardware setup.

{\flushleft \bf Challenge-sensitive performance.}
For analyzing attribute-sensitive performance of our approach against other methods, we show the quantitative comparisons in Table~\ref{tb::A2}. We evaluate our method with ten attributes (i.e., BSO, SSO, MSO, LI, CB, CIB, SA, TC, IC, OF) on the VT1000 dataset. Notice that our method outperforms other RGB-T methods on most of the challenges except BSO and LI subsets. On BSO (Big Salient Object) subset, our result ranks fourth is 4.2$\%$ less than RBD\cite{Zhu2014Saliency} in F-measure. RBD achieves better performance for introducing boundary connectivity that characterizes the spatial layout of image regions with image boundaries. On LI(Low Illumination) subset, our method ranks third is 3.2$\%$ less than FCNN\cite{Li2016DeepSaliency} in F-measure. That is because not all of the thermal infrared images are complementary to RGB images under low illumination conditions. However, FCNN obtains highest F-measure, as it can be trained with various data to handle several challenges such as low illumination, etc. In contrast, we only use features trained offline on the datasets used for other tasks such as~\cite{ZhangAJLL18}.

\subsection{Evaluation on the VT821 Dataset}
For further prove the effectiveness of the proposed approach, we also conduct experiments on the public benchmark dataset, i.e., VT821~\cite{Li2017A}. Table~\ref{tab:freq1} and Fig.~\ref{fig::pr1} present comparison results of our methods with other state-of-the-art methods, and the results further show that our method significantly outperforms other RGB-T methods (including some deep learning methods). Our method significantly outperforms the latest RGB-T method MTMR~\cite{Li2017A}, achieving 6.4$\%$ gain in F-measure over it. In particular, our method outperforms 9.1$\%$ over FCNN and 10.5$\%$ over DSS in F-measure. Through it, we could justify the effectiveness of the collaborative graph learning based on different modalities.
We also perform evaluation with different attributes (i.e., BSO, SSO, MSO, LI, CB, CIB, SA, TC, IC, OF) in the VT821 dataset, as shown in Table~\ref{tb::A1}. It is easy to see that our approach outperforms other RGB-T methods on most of the challenges, except BSO and CIB subsets, based on which our method ranks second, while RBD\cite{Zhu2014Saliency} and FCNN\cite{Li2016DeepSaliency} obtain the best, respectively. We have given some introductions of the two methods in the above. In future work, we will take appearance consistency and background prior knowledge into consideration to improve the robustness of our method.
\begin{figure}[t]
\includegraphics[width=0.4\textwidth]{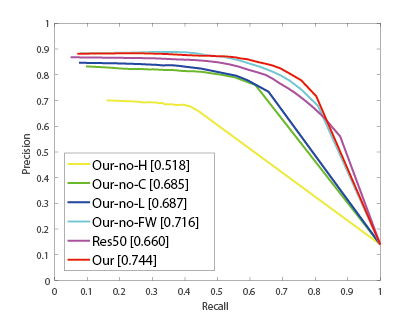}
\caption{Evaluation results of the proposed approach using different convolutional layers from the FCN-32S~\cite{Long2015Fully}, ResNet-50(Res50)~\cite{He2016Deep}. The representative score of F-measure is presented in the legend.}
\label{fig::component_feature}
\end{figure}

\begin{figure}[t]
\includegraphics[width=0.4\textwidth]{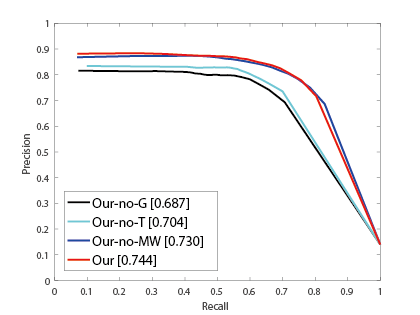}
\caption{Evaluation results of the proposed approach with its variants on the VT821 dataset. The representative score of F-measure is presented in the legend.}
\label{fig::component_modality}
\end{figure}

\subsection{Ablation Study}
To justify the effectiveness of main components of the proposed approach, we present experimental results of feature analysis and modality analysis induced from the proposed algorithm on the VT821 dataset.
{\flushleft \bf Feature Analysis.}
To perform feature analysis, we implement 5 variants, and they are: 1) Our-no-H, that removes high-level deep features in graph learning, 2) Our-no-C, that removes handcrafted color features in graph learning, and 3) Our-no-L, that removes low-level deep features in graph learning. Overall, all features contribute to boosting the final performance, and high-level deep features are most important as they encode object semantics and can distinguish objects from background effectively. 4) Our-no-FW, that removes the feature weights in graph learning. Compared with the results of Our-no-FW, our results proved the effectiveness of the introduced feature weights, which are helpful to achieve adaptive incorporation of different features information. 5) Res50, We further implement an alternative baseline method (Res50) using the first and last convolutional layer of the ResNet-50~\cite{He2016Deep}. However, we find that the result of this method is not very good. We also evaluate the performance of other layers of ResNet, but do not gain performance. This is because ResNet uses a skip connection to combine different layers, more proofs can be obtained from~\cite{Chao2017Robust}. It is worth noting that we extract feature by FCN-32S(VGG-19)~\cite{Long2015Fully}. The PR curves and F-measures are presented in Fig.~\ref{fig::component_feature}.
{\flushleft \bf Modality Analysis.}
In order to verify the validity of each modality information, we implement 3 variants, 1) Our-no-G, that removes the RGB information in our feature presentation. 2) Our-no-T, that removes the thermal information in our feature presentation. 3) Our-no-MW, that removes modality weights in graph learning. The results demonstrate the effectiveness of the introduced modality weights, which are helpful to achieve adaptive incorporation of different modal information. The PR curves and F-measures are presented in Fig.~\ref{fig::component_modality}.

\begin{table*}\footnotesize
\centering
\caption{Average runtime comparison on the VT821 dataset.}\label{tab_speed}
\begin{tabular}{|l|c|c|c|c|c|c|c|c|c|c|c|c|c|c|c|}
  \hline
  ~\textbf{Method} &MR~\cite{Yang2013Saliency} & RBD~\cite{Zhu2014Saliency} & CA~\cite{Qin2015Saliency} & RRWR~\cite{Li2015Robust} &MILPS~\cite{Huang2017Salient} & FCNN~\cite{Li2016DeepSaliency} & DSS~\cite{HouCvpr2017Dss} & MTMR~\cite{Li2017A} & Ours\\
   \hline
    ~\textbf{Runtime(s)}~& 0.55 & 3.78 & 0.79 & 1.57 & 93.2 & 0.13 & 0.06 & 1.89 & 2.23 \\
    \hline
    %~Code~ & M & M & M & M & M & C++ & P & P & M & M\\
%    \hline

\end{tabular}

\end{table*}
\subsection{Runtime Comparison}
All results were calculated for RGB-T image pairs on a Windows 10 64 bit operating system running MATLAB 2016a, with a i7 4.0GHz CPU and 32GB RAM. In Table~\ref{tab_speed}, we compare the average running time on the VT821 dataset with other state-of-the-art algorithms. The proposed algorithm costs an average of 2.23s for calculating an RGB-T image pair of 480 $\times$ 640 without considering the computational cost of extracting deep features just as~\cite{ZhangAJLL18}. Our speed is not very fast, that's because the $\bf W$-subproblem and the $\bf S$-subproblem have the calculation of the inversion operation of a matrix, which are time-consuming. In the future, to handle this problem, we will adopt a linearized operation~\cite{Lin2011Linearized} to avoid matrix inversion for efficiency.

\subsection{Failure Cases}
\begin{figure}[t]
\centering
\includegraphics[width=0.45\textwidth]{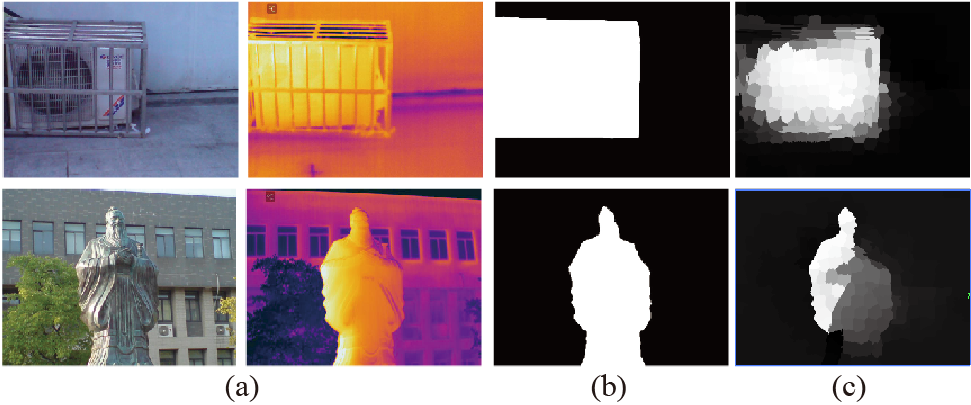}
\caption{Failure cases. (a) Input RGB-T image pairs. (b) Ground truth. (c) Saliency maps.}
\label{fig::failure_case}
\end{figure}

In this work, we utilize the collaborative graph learning approach for RGB-T saliency detection. It is proved to be effective in most cases of RGB-T saliency detection. However, when objects cross image boundary or have similar appearances with the background in both of modalities, our algorithm cannot make the salient region keeping a good contour. The main reason is when RGB and thermal data are collected in such a complicated situation, they cannot play their complementary role, and we use the boundary nodes as the initial background seeds, so the region that closes to the boundary is difficult to be detected. We also present the unsatisfying results generated by our method, as shown in Fig.~\ref{fig::failure_case}.

\section{Conclusion}
In this paper, we have proposed the collaborative graph learning approach for RGB-T saliency detection. We pose saliency detection to a graph learning problem, in which we jointly learn graph structure, edge weights (i.e., graph affinity), node weights (i.e., saliency values), modality weights and feature weights in a unified optimization framework. To facilitate performance evaluation of different algorithms, we have contributed a comprehensive dataset for the purpose of RGB-T saliency detection. Extensive experiments have demonstrated the effectiveness of the proposed approach. In future work, we will expand the dataset for large-scale evaluation of different deep learning methods, and investigate more prior knowledge to improve the robustness of our model.

%\section*{Acknowledgment}


\ifCLASSOPTIONcaptionsoff
  \newpage
\fi

\bibliographystyle{IEEEtran}
\bibliography{mybibfile}

\begin{IEEEbiography}[{\includegraphics[width=1in,height=1.25in,clip,keepaspectratio]{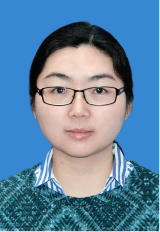}}]
{Zhengzheng Tu}received the M.S. and Ph.D. degrees from the School of Computer Science and Technology, Anhui University, Hefei, China, in 2007 and 2015, respectively. Her current research interests include computer vision, deep learning.
\end{IEEEbiography}

\begin{IEEEbiography}[{\includegraphics[width=1in,height=1.25in,clip,keepaspectratio]{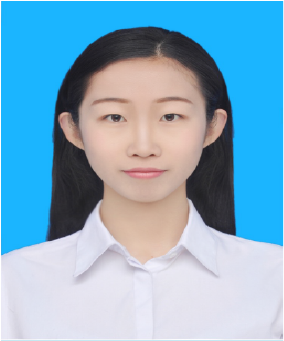}}]
{Tian Xia}received the B.S. degree in Huainan Normal University, Anhui,in 2017. She is pursuing M.S. degree in Anhui University, Hefei, China. Her current research interests include saliency detection, deep learning.
\end{IEEEbiography}

\begin{IEEEbiography}[{\includegraphics[width=1in,height=1.25in,clip,keepaspectratio]{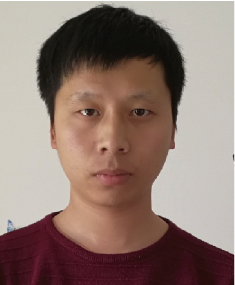}}]
{Chenglong Li} received the M.S. and Ph.D. degrees from the School of Computer Science and Technology, Anhui University, Hefei, China, in 2013 and 2016, respectively. From 2014 to 2015, he worked as a Visiting Student with the School of Data and Computer Science, Sun Yat-sen University, Guangzhou,
China. He was a postdoctoral research fellow at
the Center for Research on Intelligent Perception
and Computing (CRIPAC), National Laboratory of
Pattern Recognition (NLPR), Institute of Automation, Chinese Academy of
Sciences (CASIA), China. He is currently an Associate Professor at the School
of Computer Science and Technology, Anhui University. His research interests include computer vision and deep learning.  He was a recipient of the ACM Hefei Doctoral
Dissertation Award in 2016.
\end{IEEEbiography}

\begin{IEEEbiography}[{\includegraphics[width=1in,height=1.25in,clip,keepaspectratio]{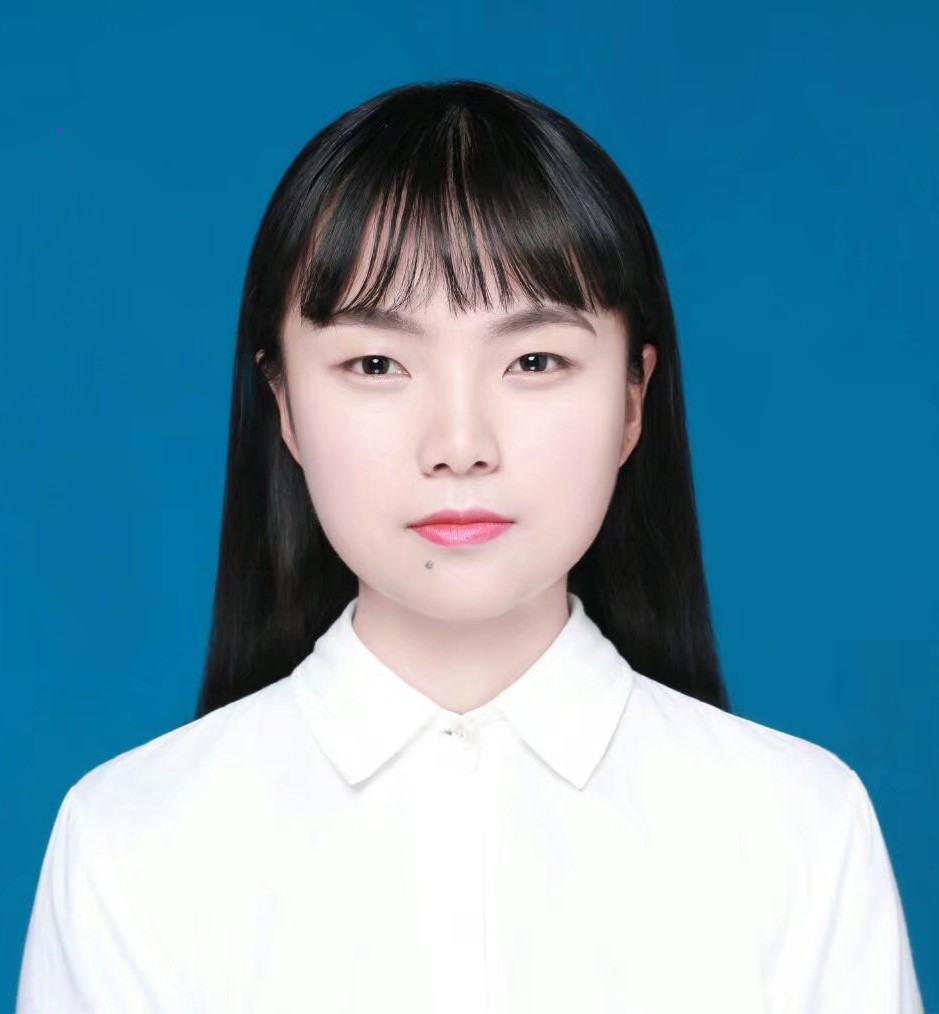}}]
{Xiaoxiao Wang} received the B.S. degree from Wannan Medical College, Wuhu, China, in 2018. She is currently pursuing the M.S. degree with Anhui University, Hefei, China. Her current research interests include computer vision and deep learning.

\end{IEEEbiography}

\begin{IEEEbiography}[{\includegraphics[width=1in,height=1.25in,clip,keepaspectratio]{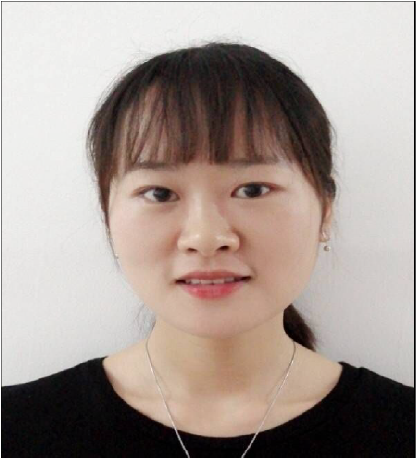}}]
{Yan Ma} received the B.S. degree in Fuyang Normal University, Anhui,in 2018. She is pursuing M.S. degree in Anhui University,Hefei,China.Her current research is saliency detection based on deep learning.
\end{IEEEbiography}

\begin{IEEEbiography}[{\includegraphics[width=1in,height=1.25in,clip,keepaspectratio]{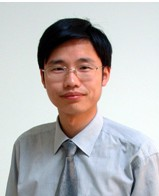}}]
{Jin Tang}received the B.Eng. degree in automation and the Ph.D. degree in computer science from Anhui University, Hefei, China, in 1999 and 2007, respectively. \\
He is currently a Professor with the School of Computer Science and Technology, Anhui University. His current research interests include computer vision, pattern recognition, machine learning and deep learning.
\end{IEEEbiography}

\end{document}